\def\year{2021}\relax
\documentclass[letterpaper]{article} 
\usepackage{aaai21}  
\usepackage{times}  
\usepackage{helvet} 
\usepackage{courier}  
\usepackage[hyphens]{url}  
\usepackage{graphicx} 
\usepackage{array}
\usepackage{colortbl}
\usepackage{makecell}
\usepackage{arydshln}
\usepackage{subfigure} 
\usepackage{booktabs}
\usepackage{amsmath}
\usepackage{algorithm}  
\usepackage[switch]{lineno} 
\usepackage[noend]{algpseudocode}
\usepackage{varwidth}
\usepackage{multirow}
\usepackage{amsfonts}

\usepackage{natbib}  
\usepackage{caption} 
\title{Topic-Aware Multi-turn Dialogue Modeling\footnote{Accepted by AAAI 2021}}
\author {
        Yi Xu,\textsuperscript{\rm 1,2,3}
        Hai Zhao, \textsuperscript{\rm 1,2,3,\thanks{Corresponding author. This paper was partially supported by National Key Research and Development Program of China (No. 2017YFB0304100), Key Projects of National Natural Science Foundation of China (U1836222 and 61733011), Huawei-SJTU long term AI project, Cutting-edge Machine reading comprehension and language model.}}
        Zhuosheng Zhang \textsuperscript{\rm 1,2,3} \\
}
\affiliations {
    \textsuperscript{\rm 1} Department of Computer Science and Engineering, Shanghai Jiao Tong University \\
    \textsuperscript{\rm 2} Key Laboratory of Shanghai Education Commission for Intelligent Interaction and \\ Cognitive Engineering, Shanghai Jiao Tong University, Shanghai, China \\
    \textsuperscript{\rm 3} MoE Key Lab of Artificial Intelligence, AI Institute, Shanghai Jiao Tong University\\
    
    {{\tt xuyi\_2019@sjtu.edu.cn, zhaohai@cs.sjtu.edu.cn, zhangzs@sjtu.edu.cn}}
}

\begin{document}

\maketitle
\begin{abstract}
In the retrieval-based multi-turn dialogue modeling, it remains a challenge to select the most appropriate response according to extracting salient features in context utterances. As a conversation goes on, topic shift at discourse-level naturally happens through the continuous multi-turn dialogue context. However, all known retrieval-based systems are satisfied with exploiting local topic words for context utterance representation but fail to capture such essential global topic-aware clues at discourse-level. 
Instead of taking topic-agnostic $n$-gram utterance as processing unit for matching purpose in existing systems, this paper presents a novel topic-aware solution for multi-turn dialogue modeling, which segments and extracts topic-aware utterances in an unsupervised way, so that the resulted model is capable of capturing salient topic shift at discourse-level in need and thus effectively track topic flow during multi-turn conversation. Our topic-aware modeling is implemented by a newly proposed unsupervised topic-aware segmentation algorithm and Topic-Aware Dual-attention Matching (TADAM) Network, which 
matches each topic segment with the response in a dual cross-attention way. Experimental results on three public datasets show TADAM can outperform the state-of-the-art method, especially by 3.3\% on E-commerce dataset that has an obvious topic shift.
\end{abstract}

\section{Introduction}
There are generally two ways to build a dialogue system, generation-based and retrieval-based. The former views conversation as a generation problem \citep{10.5555/3298023.3298055, AAAI1714571,serban2017hierarchical,zhou2017mechanism,wu2018neural}, 
while the latter usually consists of a retrieval and matching process \citep{zhou2016multi,wu-etal-2017-sequential,zhou2018multi,zhu-etal-2018-lingke,zhang-etal-2018-modeling,zhang2018gaokao,tao2019multi,gu2019interactive,zhang2020neural,zhang2019sgnet}. 



\begin{table}
\renewcommand\tabcolsep{3pt}
    \centering
    \begin{tabular}{cl}
    \toprule
           \textbf{Turns} & \multicolumn{1}{c}{\textbf{Dialogue Text}} \\
           \midrule
           Turn-1 & A: \textit{Are there any discounts activities recently?} \\
        Turn-2 & B: \textit{No. Our product have been cheaper than before.} \\
        Turn-3 & A: \textit{Oh.} \\
        Turn-4 & B: \textit{Hum!} \\
        Turn-5 & A: \textit{I'll buy these nuts. Can you sell me cheaper?} \\
        Turn-6 & B: \textit{You can get some coupons on the homepage. }\\
        \arrayrulecolor{red}\hline
        \arrayrulecolor{red}\hline
        Turn-7 & A: \textit{Will you give me some nut clips?} \\
        Turn-8 & B: \textit{Of course we will.} \\
        Turn-9 & A: \textit{How many clips will you give?} \\
        \arrayrulecolor{black} \bottomrule
    \end{tabular}
    \caption{ \label{tab:dialogue_case}A case in \citet{yuan2019multi} from E-commerce Corpus. The topic has changed after “Turn-6”.  
    }
\end{table}

The retrieval-based response selection task
is aimed to select a most suitable response from a collection of candidate answers according to a dialogue history.
Early studies concatenate context together and then match with each candidate response \citep{lowe-etal-2015-ubuntu,kadlec2015improved,10.1145/2911451.2911542,tan2015lstm,10.5555/3060832.3061030,wang-jiang-2016-learning}. Recently, most works turn to explore the interaction between the response and each utterance, which is then fused for a final matching score \citep{zhou2016multi,wu-etal-2017-sequential,zhang-etal-2018-modeling,zhou2018emotional,zhou2018multi,tao2019multi,yuan2019multi}.  To be specific, recent models follow an architecture consisting of two parts: Encoder and Matching Network. The former either encodes each context utterance  independently with traditional RNNs \citep{cho2014learning}, or models the whole context with a pre-trained contextualized language model, and then splits out utterances for further matching \citep{zhu2020dual,zhang2019dcmn+}. The latter matching modules vary in previous works, for example, 
MSN \citep{yuan2019multi} introduces a multi-hop selector which selects relevant utterances to reduce matching noise. 


In multi-turn dialogue modeling, topic-aware clues have been more or less considered as it is certain that a long enough multi-turn dialogue may have multiple topics as the conversation goes on and topic shift naturally happens by all means. Table \ref{tab:dialogue_case} shows an example that there is topic change after Turn-6. Even though existing work like \citet{yuan2019multi} select semantic-relevant information or like \citet{wu2018response} use topic information at word-level for better matching, all known systems keep using topic-agnostic or topic-mixed $n$-gram utterances as a whole for matching context. 
Instead, this work explicitly extracts topic segments from the dialogue history 
as basic units for further matching, which is capable of tracking global topic flow throughout the entire multi-turn dialogue history at discourse-level.
The proposed topic-aware modeling method groups topic-similar utterances so as to capture salient information and keep robust to irrelevant contextual noise at most degree.

To implement the proposed topic-aware modeling, we present a novel response selection model, named TADAM (Topic-Aware Dual-Attention Matching) network. For encoding, segments and response are concatenated and fed into an encoder which usually adopts a pre-trained language model, and then separated for further matching. 
For the matching network, we use selectors to weight segments based on relevance with the response in both word and segment levels, and then use cross-attention in a dual way for the multi-turn matching. Especially, we emphasize the last segment, which is the closest to the answer. Finally, we fuse these matching vectors for a final score. 

As to our best knowledge, this is the first attempt of handling multi-turn dialogue model in a topic-aware segmentation way.
Thus, for lack of multi-turn dialogue datasets with labels for dialogue topic boundaries, we label or splice two datasets in Chinese and English, respectively\footnote{Both datasets and code are available at \url{https://github.com/xyease/TADAM}}.

Our model especially fits dialogue scenes which have obvious topic shift. Experiments show our proposed model achieves state-of-the-art performance on three benchmark datasets.
\section{Related Work}
As our work concerns about topic, we have to consider how to well segment text with multiple sentences into topic-aware units. For such a purpose, previous methods vary in how they represent sentences and how they measure the lexical similarity between sentences \citep{joty2013topic}. TextTiling \citep{hearst1997texttiling} proposes pseudo-sentences and applies cosine-based lexical similarity on term frequency. Based on TextTiling, LCSeg \citep{galley2003discourse} introduces lexical chains \citep{morris1991lexical} to build vectors. To alleviate the data sparsity from term frequency vector representation, \citet{choi-etal-2001-latent} employ Latent Semantic Analysis (LSA) for representation and \citet{Song2016DialogueSS} further use word embeddings to enhance TextTiling.

For the concerned task in this paper, response selection in multi-turn dialogues, early studies conduct single-turn match, which directly concatenates all context utterances and then match with the candidate response \citep{lowe-etal-2015-ubuntu,kadlec2015improved,10.1145/2911451.2911542,tan2015lstm,10.5555/3060832.3061030,wang-jiang-2016-learning}. Recent works tend to explore relationship sequentially among all utterances, which generally follow the representation-matching-aggregation framework \citep{zhou2016multi,wu-etal-2017-sequential,zhang-etal-2018-modeling,zhou2018emotional,zhou2018multi,tao2019multi,yuan2019multi}.   In representation, there are two main approaches. One is to encode each utterance separately while the other is to encode the whole context using pre-trained contextualized language models \citep{devlin2018bert,Lan2020ALBERT:,liu2019roberta, clark2020electra, zhang2020SemBERT,zhang2021retro}, and then split out each utterance. For matching networks, there are various forms. For example, 
DAM \citep{zhou2018multi} proposes to match a response with its multi-turn context-based entirely on attention. MSN \citep{yuan2019multi} filters out irrelevant context to reduce noise. In the aggregation process, recent studies use GRU \citep{cho2014learning} with utterance vectors as input and obtain a matching score based on the last hidden state. Recently, many works such as SA-BERT \citep{Gu:2020:SABERT:3340531.3412330}, BERT-VFT \citep{whang2020domain}, DCM \citep{li58deep} have applied pre-training on the domain dataset, and get much improvement with their settings like speaker embeddings. 

As to incorporate topic issues into the dialogue system, all existing work were satisfied with introducing local topic-aware features at word-level. \citet{wu2018response} introduce two topic vectors in the matching step, which are linear combinations of topic words of the context and the response respectively. The idea of topic words are also used to generate informative responses in generation-base chatbots \citep{10.5555/3298023.3298055}. Besides, \citet{chengunrock} construct an open-domain dialogue system Gunrock,  which does 'topic' classification  for each randomly segmented text-piece and then assigns domain-specific dialogue module to generate response. However, 'topic' quoted by \citet{chengunrock} actually refers to \textit{intent}, which is very coarse across all domains, such as music, animal. When multi-turn dialogue naturally consists of multiple topics from a perspective of discourse, all known systems and existing studies ignore such essential global topic-aware clues and keep handling multi-turn dialogue in terms of topic-agnostic segments and the only adopted topic-aware information is introduced at word-level.

Different from all the previous studies, this work for the first time proposes a novel topic-aware solution which explicitly extracts topic segments from multi-turn dialogue history and thus is capable of globally handling topic-aware clues at discourse-level. 
Our topic-aware model accords with realistic dialogue scenes where topic shift is a common fact as a conversation goes on.

%
\section{Topic-Aware Multi-turn Dialogue Modeling}
Our model consists of two parts: (1)Topic-aware Segmentation, which segments the multi-turn dialogue and extracts topic-aware utterances in an unsupervised way.
(2) TADAM Network, which matches candidate responses with topic-aware segments to select a most appropriate one. 
\subsection{Topic-aware Segmentation}
\begin{algorithm}[t]
\caption{Topic-aware Segmentation Algorithm}
\label{DTSG}
\begin{algorithmic}[1] 
\Require
Dialogue $D=\{{u_1},{u_2},...,{u_{n}}\}$
\Ensure 
Topic segment list $S$
\State $S$=[ ], start index $i=1$
\While {$i \le  n $}
\State ${l} = {u_{i - d}} \oplus  \cdots  \oplus {u_{i- 1}}$
\State $j=1, {c_0}=``",$
\While {$i+j \le n+1$ \textbf{and} $j \le R$}
\State $c_j = {c_{j-1}} \oplus {u_{i+j-1}}$
\If {$j$ mod $K$ == 0}
\State $r_j = {u_{i + j }} \oplus  \cdots  \oplus {u_{i + j + d-1}}$
\State \begin{varwidth}[t]{\linewidth}
$cos{t_{{c_j}}} = \max($ 
${\rm sim}(E({c_j}),E(l))$, \\
${\rm sim}(E({c_j}),E({r_j})))$
\end{varwidth}
\EndIf
\State $j+=1$
\EndWhile
\State $c_j^ *  = \mathop {\min }\limits_{{c_j}} cos{t_{{c_j}}}$
\State $S.{\rm append}(c_j^ *), i=\left| {c_j^ * } \right|+1$
\EndWhile
\Return $S$
\end{algorithmic}
\end{algorithm}
Considering that a continuous multi-turn dialogue history $D$ with $n$ utterances, where $D=\{u_1,...,u_{n}\}$, we design an unsupervised topic-aware segmentation algorithm to determine segmentation points from intervals between $u_i$ as shown in Algorithm \ref{DTSG}, where $E(\cdot)$ is to encode text sequence and more details is in Section \ref{sec: seg_result}.

The proposed segmentation algorithm greedily checks each interval between two adjacent utterances to determine a segmentation that lets two resulted segments mostly differ. In detail, for each candidate interval, we concatenate utterances from the previous segmentation point as center segment $c$ and set a sliding window $d$ to get left and right segments $l,r$. We expect to get segment $c$, which is the least similar to $l,r$. 

Besides, considering the number of utterances in each topic segment, we set a maximum range $R$ within which segmentation may happen from last point. In addition, to avoid yielding too many fragments, interval check is actually performed skipping every $K$ ones. We also set a threshold $\theta_{cost}$ in Algorithm \ref{DTSG}, i.e., if $cost>\theta_{cost}$, then there is no segmentation so as to further control fragmentation.
\subsection{TADAM Network}
After topic-aware segmentation, dialogue history is segmented into topic segments, which are then as context input of TADAM to keep self-informative and robust to irrelevant contextual information.

We denote the dataset as $D = \{ {(C,R,Y)_k}\} _{k = 1}^N$, where $C$ is dialogue context, and $R$ is the candidate response, and $Y \in \{ 0,1\}$ is the label indicating whether $R$ is a proper response for $C$. Besides, $C = \{ {S_1},{\rm{ }}...,{S_n}\}$ and ${S_i},1 \le i \le n$ is the $i$-th topic segment in context $C$.

\begin{figure*}[ht]
\centering
\includegraphics[scale=1.06]{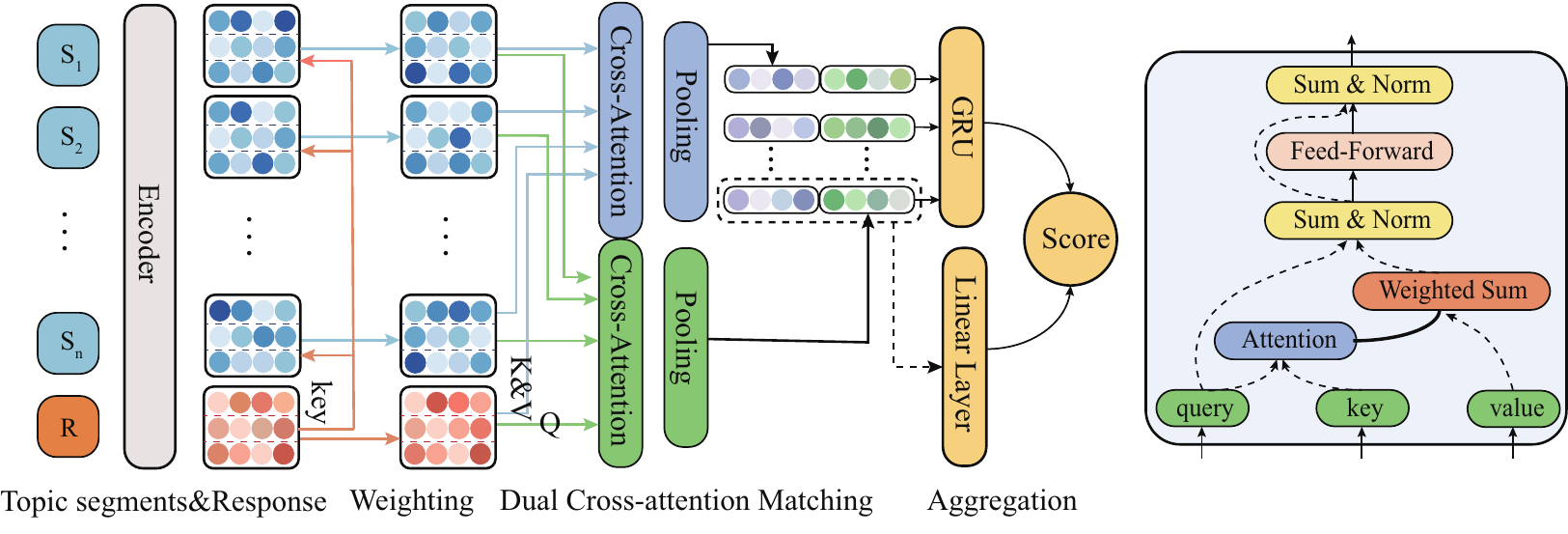}
\caption{Architecture of our TADAM network (left) and Attentive Module (right).}
\label{fig:Module}
\end{figure*}

Figure \ref{fig:Module} shows the architecture of TADAM. For encoding, all segments and response are concatenated as an input sequence for an encoder and then separated out according to position information. For matching, we first use response to weight each segment at both word and segment levels in order to pay more attention to those which are more related. Then, the response is used to match each weighted segment in a dual cross-attention way for deep matching of each pair of segment and response, which outputs a matching vector for each segment. Those matching vectors are fed into GRU for a single one. Ultimately, we fuse the single vector with the matching vector of the last segment for a score.
\paragraph{Encoding and Separation}
We take well pre-trained language model as our encoder component. Following the encoding manner of pre-trained language models such as BERT and ALBERT, we concatenate all segments $\{ {S_{i}}\} _{i=1}^n$ and response $R$ with special tokens: $X =  \left[ \texttt{CLS} \right]{S_1}\left[ \texttt{SEP} \right].{\rm{ }}.{\rm{ }}.\left[ \texttt{SEP} \right]{S_n}\left[ \texttt{SEP} \right]R\left[ \texttt{SEP} \right]$, which is fed into the encoder. Then we split the segments and response according to position information. Because the number of segments $n$ is different among all contexts, and the length of each segment varies, we set the maximum number of segments in each context as $T$, and  the maximum length of segments as $L$. Thus after separation, we can get context representation $C \in {\mathbb{R}^{T \times L \times d}}$, where $d$ is dimension of each token in the encoder output. Response representation $R \in {\mathbb{R}^{L \times d}}$ and $C$ contains $T$ segment representation $\{ {S_i} \in {\mathbb{R}^{L \times d}}\} _{i = 1}^T$. More encoding modes are discussed in Section \ref{sec:inputmode}.
\paragraph{Segment Weighting}
\label{sec: Segment Weight}
\citet{yuan2019multi} use the last $k$ utterances of context to select relevant and meaningful context utterances at both word and utterance granularities. Considering our process unit is segments, which are much longer than single utterances, it will lead to information scarcity if we delete the whole unselected segments. Instead, we use the candidate response as a key utterance to give each segment a weight at both word and segment granularities.

\noindent $\bullet$ \textbf{Word-level Weights:}
At the word level, we build a matching feature map between each segment $S_i$ and response $R$, which is formulated as:
\begin{align*}
& {M^1} ={\rm transfer}(C,W,R), \\
&{M} = \frac{1}{{\sqrt d }}{\rm tanh}({M^1})V, \\
& {M^{pool}} = [\mathop {\max M}\limits_{\dim  = 2} ,\mathop {\max M}\limits_{\dim  = 3} ],
\end{align*}
where $M_{xyuv}^1 = \sum\limits_k {{C_{xyk}}{W_{kkv}}{R_{uk}}}$. $W \in \mathbb{R}^{d \times d \times h}, V \in \mathbb{R}^h$ are learnable parameters, and ${M^1} \in {\mathbb{R}^{T \times L \times L \times h}}$, $M \in \mathbb{R}^{T \times L \times L}$, ${M^{pool}} \in \mathbb{R}^{T \times 2L}$. The matching map $M$ is max-pooled in both row and column. 
$M^{pool}$ indicates the relevance between response $R$ and $T$ segments at the word level. Then we transfer matching features $M^{pool}$ into weights for $T$ segments through a linear layer:
\begin{equation*}
{s_1} = {\rm softmax} (M^{pool}W' + b),
\end{equation*}
where $s_1 \in \mathbb{R}^{T}$ is segment weights at word level, and $W' \in \mathbb{R}^{2L},b \in \mathbb{R}^T$ are learnable parameters.

\noindent $\bullet$ \textbf{Segment-level Weights:}
At the segment level, we build segment representation by mean-pooling token vectors so that the segment is represented by a vector. Then we use cosine similarity with the response $R$ to get a weight for each segment:
\begin{align*}
&C'={\rm mean}{(C)_{\dim  = 2}}, \\
&{s_2} = \cos (C',R),
\end{align*}
where $C' \in \mathbb{R}^{T\times d}$, $s_2 \in \mathbb{R}^{T}$. $s_2$ is segment weights at the segment level, which catches the overall semantic similarity between the response and segment.

\noindent $\bullet$ \textbf{Combination:}
In order to fuse both two levels, we set a hyper-parameter $\alpha$ and sum up $s_1,s_2$. Then we multiply $s$ and $C$ to give segments different degree of relevance based on the response:
\begin{align*}
&s=\alpha s_1+(1-\alpha)s_2, \\
&{\tilde C}=s \odot C,
\end{align*}
where $s \in \mathbb{R}^{T}$, ${\tilde C} \in \mathbb{R}^{T\times L \times d}$.
\paragraph{Dual Cross-attention Matching}
We also use the Attentive Module in DAM \citep{zhou2018multi} which is a unit of transformer \citep{vaswani2017attention} to encode the interaction between two sequences. Here we apply cross-attention to the segment and response in a dual way, which is fused for further process.

\noindent $\bullet$ \textbf{Attentive Module:}
The architecture of Attentive Module is shown in Figure \ref{fig:Module}, which takes three sequences as input: query sequence $Q\in \mathbb{R}^{n_q\times d}$, key sequence $K\in \mathbb{R}^{n_k\times d}$ and value sequence $V\in \mathbb{R}^{n_v\times d}$, where $n_q,n_k,n_v$ are the number of tokens respectively, and $d$ is embedding dimension.


%

Attentive Module first takes each word in the query sentence to attend to words in the key sentence via Scaled Dot-Product Attention \citep{vaswani2017attention}, then applies those attention results upon the value sentence, which is defined as:
\begin{equation*}
{V_{att}} = {\rm softmax} (\frac{{Q{K^T}}}{{\sqrt d }})V,
\end{equation*}
where $V_{att} \in \mathbb{R}^{n_q\times d}$. Then layer normalization \citep{ba2016layer} takes $V_{att}$ as input and we denote the output as $V'_{att} \in \mathbb{R}^{n_q\times d}$, which is then fed into a feed-forward network FFN with RELU \citep{lecun2015deep} activation:
\begin{equation*}
FFN(V'_{att}) ={\rm max} (0,{V'_{att}}{W_1} + {b_1}){W_2} + {b_2},
\end{equation*}
where $W_1, b_1, W_2, b_2$ are learnable parameters and $FFN(V'_{att}) \in \mathbb{R}^{n_q\times d}$. We denote the whole Attentive Module as:
 \begin{equation*}
 {\rm AttentiveModule}(Q,K,V),
 \end{equation*}
 
\noindent $\bullet$ \textbf{Dual Cross-Attention Matching:}
For each segment ${\tilde S_i} \in \mathbb{R}^{L \times d}$ in weighted context ${\tilde C} \in \mathbb{R}^{T\times L \times d}$, we use cross-attention to build $S_i^{cross}$, each token of which reflects the relevance with response $R$, and $R_i^{cross}$ about relevance with ${\tilde S_i}$:
\begin{align*}
S_i^{cross}&={\rm AttentiveModule}({\tilde{S_i}},R,R),\\
R_i^{cross}&={\rm  AttentiveModule}(R,{\tilde{S_i}},{\tilde{S_i}}),
\end{align*}
where $S_i^{cross}, R_i^{cross} \in \mathbb{R}^{L \times d}$. All the $S_i^{cross}, R_i^{cross}$ construct $C_S^{cross} \in \mathbb{R}^{T \times 	L \times d}$, $C_R^{cross} \in \mathbb{R}^{T \times L \times d}$, which are then mean-pooled and concatenated as $C^{cross}$ (other pooling and fusion methods are discussed in Section \ref{sec:fusion}).
\begin{align}
\label{euq:C_S_1} &C_S^{cross}=\{ S_i^{cross}\} _{i = 1}^T, \\
\label{euq:C_R_1} &C_R^{cross}= \{ R_i^{cross}\} _{i = 1}^T,\\
\label{euq:C}&{C^{cross}} = [\mathop {{ \rm mean}(C_S^{cross})}\limits_{\dim  = 2} ,\mathop {{\rm mean}(C_R^{cross})}\limits_{\dim  = 2} ],
\end{align}
where $C^{cross} \in \mathbb{R}^{T  \times 2d}$ indicates the matching features for each segment based on the candidate response.
\paragraph{Aggregation}
\label{sec: Aggregation}
We use GRU to model the relation of segments as SMN \citep{wu-etal-2017-sequential}. Considering the last segment which is the closest to the response may contain more relevant information, we add $C_T^{cross}$, which is the dual matching between the last segment $S_T^{cross}$ and the response $R_T^{cross}$. Furthermore, we add a linear layer to $C_T^{cross}$:
\begin{align}
&\hat H = {\rm GRU}({C^{cross}})\nonumber,  \\ 
\label{euq:C_T}&C_T^{cross} = [\mathop {{\rm mean}(S_T^{crose})}\limits_{\dim  = 1} ,\mathop {{\rm mean}(R_T^{crose})}\limits_{\dim  = 1} ],
\\
&\hat C_T^{cross}=W_3C_T^{cross}+b_3 \nonumber,
\end{align}
where $\hat H, C_T^{cross}, \hat C_T^{cross} \in \mathbb{R}^{2d}$.  $W_3 \in  \mathbb{R}^{2d \times 2d}, b_3 \in \mathbb{R}^{2d}$ are learnable parameters. Finally, we get the score for context $C$ and its candidate response $r$ using a linear layer, where $W_4 \in  \mathbb{R}^{4d}, b_4 \in \mathbb{R}$ are learnable parameters (other fusion methods are discussed in Section \ref{sec:fusion}):
\begin{equation}
\label{euq:score}score = {\rm sigmoid}({W^T_4}[\hat H,\hat C_T^{cross}] + {b_4}).
\end{equation}
\section{Experiments}


\subsection{Dataset}
For topic-aware segmentation, our proposed algorithm is evaluated in two newly built datasets as it is the first attempt for topic-aware multi-turn dialogues. \textbf{For Chinese}, we annotate a dataset including 505 phone records of customer service on banking consultation. \textbf{For English}, we build dataset including 711 dialogues by joining dialogues from existing multi-turn dialogue datasets: MultiWOZ Corpus\footnote{https://doi.org/10.17863/CAM.41572} \citep{budzianowski-etal-2018-multiwoz} and Stanford Dialog Dataset \citep{eric-etal-2017-key}, where each dialogue is about one topic.

For response selection, TADAM is tested on three widely used public datasets.
(1) \textbf{Ubuntu Corpus} \citep{lowe-etal-2015-ubuntu}: consists of English multi-turn conversations about technical support collected from chat logs of the Ubuntu forum. (2) \textbf{Douban Corpus} \citep{wu-etal-2017-sequential}: consists of multi-turn conversations from the Douban group
, which is a popular social networking service in China. 
(3) \textbf{E-commerce Corpus} \citep{zhang-etal-2018-modeling}: includes conversations between customers and shopkeepers from the largest e-commerce platform Taobao 
in China. The E-commerce Corpus has an obvious topic shift, including commodity consultation, logistics express, recommendation, and chitchat.


\subsection{Evaluation}
For topic-aware segmentation, we adopt three metrics: (1) $MAE$ adopted by \citep{ijcai2018-612}, is defined as $\frac{1}{{\left| T \right|}}\sum\nolimits_{D \in T} {\left| {{N_{pred}}\left( D \right) - {N_{ref}}\left( D \right)} \right|} $, where $T$ is a dialogue, and ${N_{pred}}\left( D \right),{N_{ref}}\left( D \right)$ denote the prediction and reference number of segments in dialogue $T$. (2) WindowDiff ($WD$) adopted from \citep{pevzner2002critique}. $WD$ moves a short window through the dialogue from beginning to end, and if the number of segmentation in prediction and reference is not identical, a penalty of 1 is added. 
The window size in our experiments is set to 4, and we report the mean $WD$ for all experiments. (3) $F_1$ score is the harmonic average of recall and precision of segmentation points.

For response selection, we use the same metric $R_n@k$ as previous works, which selects $k$ best matchable candidate responses among $n$ and calculates the recall of the true ones. Besides, because Douban corpus has more than one correct candidate response, we also use MAP (Mean Average Precision), MRR (Mean Reciprocal Rank), and Precision-at-one P@1 as previous works.

\subsection{Experimental Settings}
For both tasks, we use pre-trained BERT\footnote{https://github.com/huggingface/transformers} \citep{devlin2018bert} (bert-base-uncased \& bert-base-chinese) as encoder. 

For topic-aware segmentation: In both datasets, we set range $R=8$, jump step $K=2$ (value of 1 will lead to fragmentation), window size $d=2$ and threshold $\theta_{cost}$=0.6. For TextTiling, the length of a pseudo sentence is set to 20 in the Chinese dataset and 10 in the English dataset, which is close to the mean length of utterances in both datasets. Window size and block size for TextTiling are all set to 6.

For response selection: We apply topic-aware segmentation algorithm to Ubuntu, Douban and E-commerce with range $R=2, 2, 6$ after trying different values. As to our model, the max input sequence length is set to 350 after WordPiece tokenization and the max number of segments is 10. We set the learning rate as 2e-5 using BertAdam with a warmup proportion of 10\%. Our model is trained
with batch size of \{20,32,20\} and epoch of \{3,3,4\} for Ubuntu, Douban and E-commerce.  Besides, the $\alpha$ of word-level weights is set to 0.5. As for the baseline BERT, we finetuned it. The epoch is set as \{3, 2, 3\} for three datasets, and other settings are the same as our model. 
\subsection{Results}
\label{sec: seg_result}
\textbf{Topic-aware segmentation:} To evaluate the performance of our method, we compare it with a typical text segmentation algorithm, TextTiling \citep{hearst1997texttiling}, which is a classic text segmentation algorithm using term frequency vectors to represent text. Besides, TextTiling+Embedding \citep{Song2016DialogueSS} applies word embedding to compute similarity between texts. Therefore, we also compare TextTiling with our segmentation algorithm in three representation methods: BERT$_{CLS}$, which uses "$\left[ {CLS} \right]$" embedding to directly encode the entire text, and BERT$_{mean}$, GloVe: just use mean vector of all words in the text.
 \begin{table}[!ht]
   \footnotesize
  \renewcommand\tabcolsep{1.6pt}
            \centering
\begin{tabular}{lccccccc}
\toprule 
\multirow{2}*{\textbf{Method}}& \multicolumn{3}{c}{Chinese}&~&\multicolumn{3}{c}{English} \\ 
\cmidrule(lr){2-4}
\cmidrule(lr){6-8}
 ~& \textbf{$MAE$}& \textbf{~$WD$} & \textbf{~$F_1$}&~& \textbf{$MAE$}& \textbf{~$WD$} & \textbf{~$F_1$} \\ 
\midrule
TextTiling & 1.90&0.45  & 0.52&~&10.08& 0.83&0.34 \\
TextTiling+GloVe& 2.0&0.45  & 0.52&~&6.38& 0.75&0.33\\

TextTiling+BERT$_{mean}$& 6.50&0.60 & 0.45&~&9.64& 0.81&0.32\\
TextTiling+BERT$_{CLS}$& 6.51&0.60 & 0.45&~&9.78& 0.82&0.33\\
Our algo.+ GloVe&3.83&0.61&0.48&~&3.48&0.59&0.56\\
Our algo.+BERT$_{mean}$&2.95&0.52&0.51&~&2.98&\textbf{0.52}&\textbf{0.61}\\
Our algo.+BERT$_{CLS}$ &\textbf{ 0.79} &\textbf{0.34} &\textbf{0.61}&~ & \textbf{1.04}&0.54&0.44 \\
\bottomrule
\end{tabular}
\caption{\label{tab: seg_results} Topic-aware segmentation results.}
   \end{table}

\begin{table*}[ht]
\renewcommand\tabcolsep{0.8pt}
\footnotesize
\centering
\begin{tabular}{lcccccccccccc}
\toprule 
\multirow{2}*{\textbf{Model}}& \multicolumn{3}{c}{Ubuntu}&\multicolumn{6}{c}{Douban}& \multicolumn{3}{c}{E-commerce}\\ 
~&$R_{10}@1$&$R_{10}@2$&$R_{10}@5$&MAP&MRR&P@1&$R_{10}@1$&$R_{10}@2$&$R_{10}@5$&$R_{10}@1$&$R_{10}@2$&$R_{10}@5$\\ 
\midrule 
TF-IDF \citep{lowe-etal-2015-ubuntu}& 41.0 &54.5& 70.8&33.1 &35.9& 18.0& 9.6 &17.2& 40.5&15.9&25.6&47.7 \\
RNN \citep{lowe-etal-2015-ubuntu}&40.3& 54.7& 81.9&39.0& 42.2& 20.8& 11.8 &22.3& 58.9&32.5&46.3&77.5\\
CNN \citep{kadlec2015improved}&54.9& 68.4& 89.6&41.7& 44.0 &22.6& 12.1 &25.2& 64.7&32.8&51.5&79.2\\
LSTM \citep{kadlec2015improved}&63.8 &78.4& 94.9&48.5 &53.7& 32.0 &18.7& 34.3& 72.0&36.5&53.6&82.8\\
BiLSTM \citep{kadlec2015improved}&63.0 &78.0& 94.4&47.9& 51.4 &31.3& 18.4 &33.0 &71.6&35.5&52.5&82.5\\
DL2R \citep{10.1145/2911451.2911542}&62.6 &78.3& 94.4&48.8& 52.7& 33.0 &19.3 &34.2& 70.5&39.9&57.1&84.2\\
Atten-LSTM \citep{tan2015lstm}&63.3 &78.9& 94.3&49.5& 52.3 &33.1 &19.2& 32.8& 71.8&40.1&58.1&84.9\\
MV-LSTM \citep{10.5555/3060832.3061030}&65.3& 80.4 &94.6&49.8& 53.8& 34.8 &20.2 &35.1& 71.0&41.2&59.1&85.7\\
Match-LSTM \citep{wang-jiang-2016-learning}&65.3& 79.9& 94.4&50.0 &53.7& 34.5 &20.2 &34.8& 72.0&41.0&59.0&85.8\\
\midrule
Multi-View \citep{zhou2016multi}&66.2& 80.1& 95.1&50.5& 54.3& 34.2& 20.2 &35.0& 72.9&42.1&60.1&86.1\\
SMN \citep{wu-etal-2017-sequential}&72.6 &84.7& 96.1&52.9 &56.9& 39.7& 23.3& 39.6& 72.4&45.3&65.4&88.6\\
DUA \citep{zhang-etal-2018-modeling}&75.2&86.8& 96.2&55.1 &59.9& 42.1& 24.3& 42.1& 78.0&50.1&70.0&92.1\\
DAM \citep{zhou2018multi}&76.7& 87.4& 96.9&55.0& 60.1& 42.7 &25.4& 41.0& 75.7&-&-&-\\
MRFN\citep{tao2019multi}&78.6& 88.6& 97.6&57.1&61.7& 44.8& 27.6& 43.5& 78.3
&-&-&-\\     
IMN \citep{gu2019interactive}&79.4&88.9&97.4&57.0& 61.5& 44.3 &26.2& 45.2& 78.9&62.1&79.7&96.4\\
IOI \citep{tao-etal-2019-one}&79.6&89.4&97.4&57.3& 62.1& 44.4 &26.9 &45.1 &78.6&56.3&76.8&95.0\\
MSN \citep{yuan2019multi}&80.0&89.9&97.8&58.7& 63.2& \textbf{47.0} &\textbf{29.5}& 45.2 &78.8&60.6&77.0&93.7\\
\midrule
BERT&81.9&90.4&97.8&58.7&62.7&45.1&27.6&45.8&82.7&62.7&82.2&96.2\\
TADAM (Ours)&\textbf{82.1}&\textbf{90.6}&\textbf{97.8}&\textbf{59.4}&\textbf{63.3}&45.3&28.2&\textbf{47.2}&\textbf{82.8}&\textbf{66.0}&\textbf{83.4}&\textbf{97.5}\\
\bottomrule
\end{tabular}
\caption{\label{tab: sel_results} Response selection results on on Ubuntu, Douban and E-commerce datasets.}
\end{table*}
Results are shown in Table \ref{tab: seg_results}. We can find that, just except for GloVe in the Chinese dataset, our proposed segmentation algorithm for multi-turn dialogues surpasses TextTiling in all metrics. TextTiling tends to have larger $MAE$, because it ignores the number of turns in a topic round. In our following response selection part, we use our algorithm with BERT$_{CLS}$ for topic-aware segmentation.
 
\noindent \textbf{Response selection:} First, we apply the segmentation algorithm to dialogues on Ubuntu, Douban and E-commerce datasets. 
We use the smallest number larger than average length of dialogues as the initial value of R. Then different R are tried around the initial value and we select the best one.

Besides, we concatenate the context and candidate response as input for BERT as a basic sequence classification baseline. Results in Table \ref{tab: sel_results} show that our model outperforms all public works and especially gets much improvement (3.3\% in $R_{10}@1$) over the strong pre-trained contextualized language model in E-commerce dataset, which shows the effectiveness of our topic-aware models in dialogues with topic shifting scenes. 

Through observation, we find that the fact of topic shift is negligible in Ubuntu and Douban where a whole dialogue is almost about one topic. This is why improvement of the model in Douban is not as obvious as that in E-commerce with multiple topics. However, our work is not supposed to work best in all scenarios, but especially focuses on the case of topic shift which are common in the more challenging e-commerce/banking situations. Results show that it does work in specific application scenarios where topic shift is ambiguous, which right verifies the motivation of this work.

\section{Analysis}
\subsection{Ablation Studies}
In order to investigate the performance of each part of our model, we conduct serious ablation experiments from three angles and results are shown in Table \ref{tab: ablation}. First, we explore the influence of the segment weighting part (in Section \ref{sec: Segment Weight}) by moving word or segment level weights, or both of them (line 3-5). Second, in the aggregation part (in Section \ref{sec: Aggregation}), we concatenate the multi-turn matching result $\hat{H}$ and last segment matching result $\hat C_T^{cross}$ to get a score. Hence we remove either of both each time (line 6-7). Third, we do dual cross-attention matching which includes cross-attentioned segments $C_S^{cross}$ and cross-attentioned responses $C_R^{cross}$ in Equation \ref{euq:C_S_1}, \ref{euq:C_R_1} respectively, and we explore single matching methods in line 8-9.
\begin{table}[ht]
\renewcommand\tabcolsep{0pt}
\footnotesize
\centering
\begin{tabular}{lcccccc}
\toprule  \small
\multirow{2}*{\textbf{Model}}& \multicolumn{3}{c}{E-commerce}&\multicolumn{3}{c}{Ubuntu}\\ 
~&$R_{10}@1$&$R_{10}@2$&$R_{10}@5$&$R_{10}@1$&$R_{10}@2$&$R_{10}@5$\\ 
\midrule
TADAM&66.0&83.4&97.5&82.1&90.6&97.8\\
\midrule 
w/o word weights&62.0&82.2&96.3&81.9&90.6&97.8\\
w/o seg. weights&63.1&82.7&97.0&81.8&90.5&97.8\\
w/o weights&62.2&82.2&97.5&81.9&90.6&97.7\\
\midrule
w/o last seg. match&64.0&82.9&96.7&82.0&90.5&97.8\\
w/o multi-turn match&63.8&82.5&96.4&81.9&90.5&97.8\\
\midrule
single match (seg.)&62.7&83.3&97.4&81.9&90.6&97.8\\
single match (res.)&62.4&83.2&97.6&81.6&90.3&97.8\\
\bottomrule
\end{tabular}
\caption{\label{tab: ablation} Ablation study on E-commerce and Ubuntu.}
\end{table}

As shown in Table \ref{tab: ablation}, for E-commerce, removing word or segment level weights all perform worse. Besides, adding extra last segment match does make sense as the traditional multi-turn matching method with GRU. Moreover, the single match either uses segments or response as the query sequence leads to much decrease, which shows the effectiveness of our dual matching design. Results of Ubuntu are not so obvious as that of E-commerce, which can be attributed to that our work especially focuses on the case of topic shift but dialogues in Ubuntu are almost about one topic.

\subsection{Exploring Other Input Modes}
\label{sec:inputmode}
We encode the context and candidate response by concatenating cut segments and the response with special tokens, which is fed into the encoder and then split by position information. In this part, we investigate other input modes in two respects and results are shown in Table \ref{tab: input} (line 3,4).
\begin{table}[ht]
 \centering
\begin{tabular}{lccc}
\toprule 
\textbf{Model}&$R_{10}@1$&$R_{10}@2$&$R_{10}@5$\\ 
\midrule
TADAM&66.0&83.4&97.5\\
\midrule
separated segment &36.0&49.2&72.5\\
concatenated utterance&62.9&81.5&97.2\\
\midrule
max-pool (match)&63.1&83.4&97.8\\
sum (match)&63.6&83.0&96.8\\
sum (aggregation)&62.8&84.3&97.6\\
\bottomrule
\end{tabular}
\caption{\label{tab: input} Results of different input modes \& fusion methods.}
\end{table}

First, we encode each segment and response separately, indicating that the segment or response itself just focuses on its own meaning without interaction with other contexts. Results in line 3 show that this input mode causes a great loss, which means that although the segment itself can be encoded purely, it leads to information scarcity and will be more sensitive to segmentation error. 
Table \ref{tab:dialogue_case_2} shows a topic-aware segmentation case from E-commerce Corpus. We can find that our algorithm does split out topic segments, whose boundaries are very near the true ones. On the other hand, it is not ideal to encode segments and response separately. Encoding concatenated segments allows for relevant information supplement, while separated encoding tends to suffer from mixed topics caused by segmentation deviation.
\begin{table}[ht]
\renewcommand\tabcolsep{2pt}
\footnotesize
    \centering
    \begin{tabular}{|c|l|}
    \hline
           \textbf{Turns} & \multicolumn{1}{c|}{\textbf{Dialogue Text}} \\
           \hline
           Turn-1 & A: \textit{Hello.} \\
        Turn-2 & B: \textit{Excuse me, has my order been sent out?} \\
        \arrayrulecolor{blue}\hdashline
        \arrayrulecolor{blue}\hdashline
        Turn-3 & A: \textit{Please let me check.} \\
        \arrayrulecolor{red}\hline
        \arrayrulecolor{red}\hline
        Turn-4 & B: \textit{I found I didn't buy the cotton one.} \\
        Turn-5 & A: \textit{Your order has been sent out.} \\
        Turn-6 & B: \textit{It's non-woven fabric.}\\
        Turn-7 & A: \textit{Yes.} \\
        Turn-8 & B: \textit{I'd like to switch to the plant fiber.} \\
        \arrayrulecolor{blue}\hdashline
        \arrayrulecolor{blue}\hdashline
        Turn-9 & A: \textit{Ok.} \\
        \arrayrulecolor{red}\hline
        \arrayrulecolor{red}\hline
        Turn-10 & B: \textit{Please change it for me.} \\
        Turn-11 & A: \textit{Sorry, your order has been taken by the courier.} \\
        Turn-12 & B: \textit{Can you get it back?} \\
        Turn-13 & A: \textit{I'll try to intercept for you} \\
        Turn-14 & B: \textit{I'm sorry} \\
        \arrayrulecolor{blue}\hdashline
        \arrayrulecolor{blue}\hdashline
        Turn-15 & A: \textit{It doesn't matter} \\
        \arrayrulecolor{red}\hline
        \arrayrulecolor{red}\hline
        Turn-16 & B: \textit{What is the natural plant fiber?} \\
        \arrayrulecolor{black}\Xhline{0.8pt}
    \end{tabular}
    \caption{ \label{tab:dialogue_case_2}A topic-aware segmentation case from E-commerce Corpus. Red solid lines are right boundaries, and blue dotted lines are labelled by our segmentation algorithm.}
\end{table}

\begin{figure}[ht]
\centering
\includegraphics[scale=0.26]{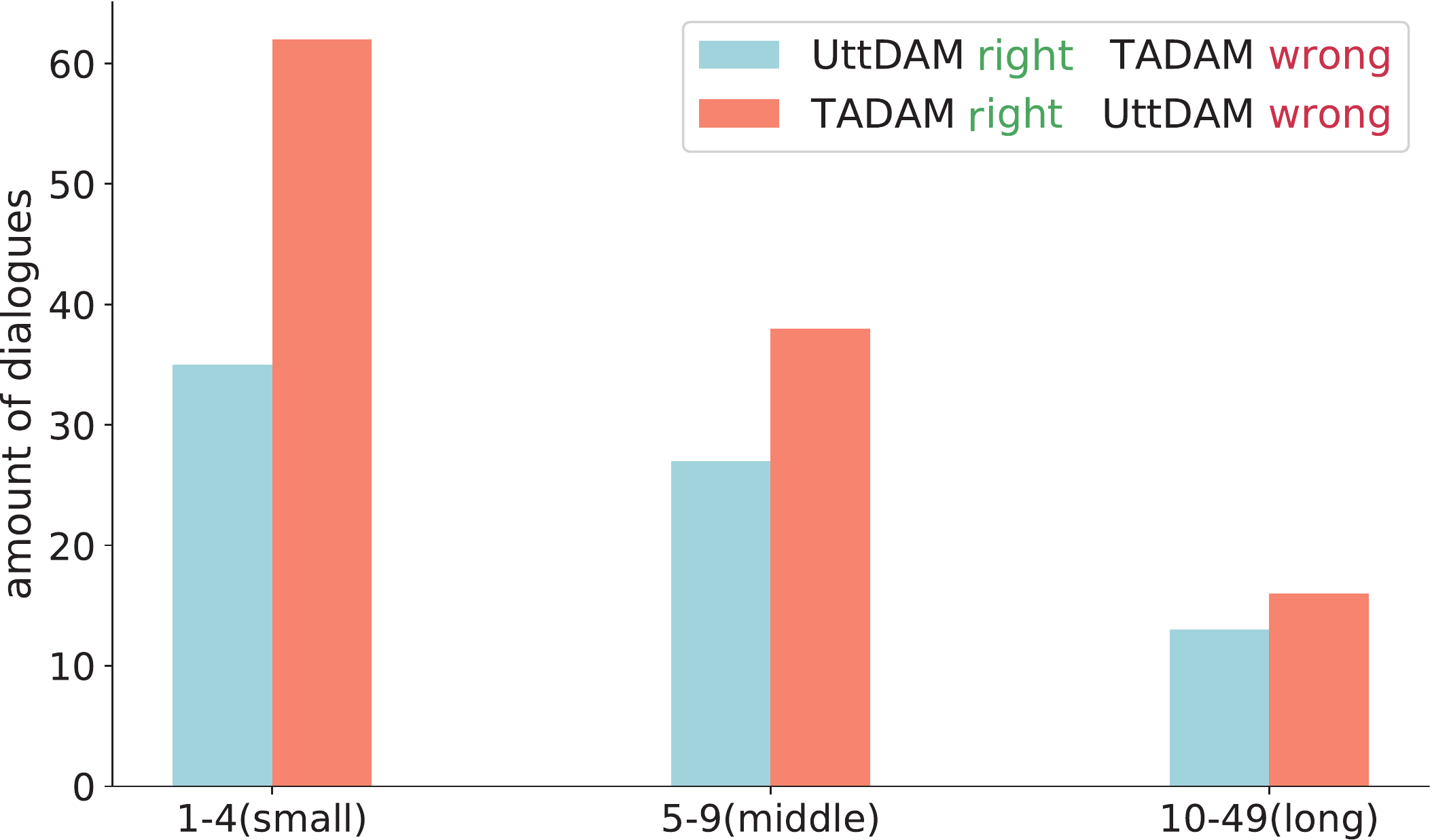}
\caption{Distribution of dialogue length}
\label{fig: dialogue len distribution}
\end{figure}

Second, we remove the segmentation part and just concatenate utterances as well as response (noted as UttDAM). Its results in line 4 of Table \ref{tab: input} also decrease much, which implies that taking the segment as a unit is more robust to the irrelevant contextual information than the utterance. 
we find that among all right justified dialogues in TADAM, 24.4\% are segmented.
Figure \ref{fig: dialogue len distribution} shows the distribution of dialogue length for both TADAM and UttDAM. We can observe that topic segment does work in dialogues of different lengths.
\subsection{Exploring Other Fusion Methods}
\label{sec:fusion}
In this section, we discuss other pooling or fusion methods in our model. Results are shown in Table \ref{tab: input} (line 5,6,7). We try max-pooling to get the representation for the whole segment with token vectors (in Equation \ref{euq:C}, \ref{euq:C_T}). Besides, we test the element-wise summation fusion method in both matching (in Equation \ref{euq:C}, \ref{euq:C_T}) and aggregation stage (in Equation \ref{euq:score}). Other fusion methods do not perform so well as TADAM, but still surpass the BERT baseline.
\subsection{Effects of Topic-aware Segmentation}
\begin{figure}[ht]
\centering
\includegraphics[scale=0.3]{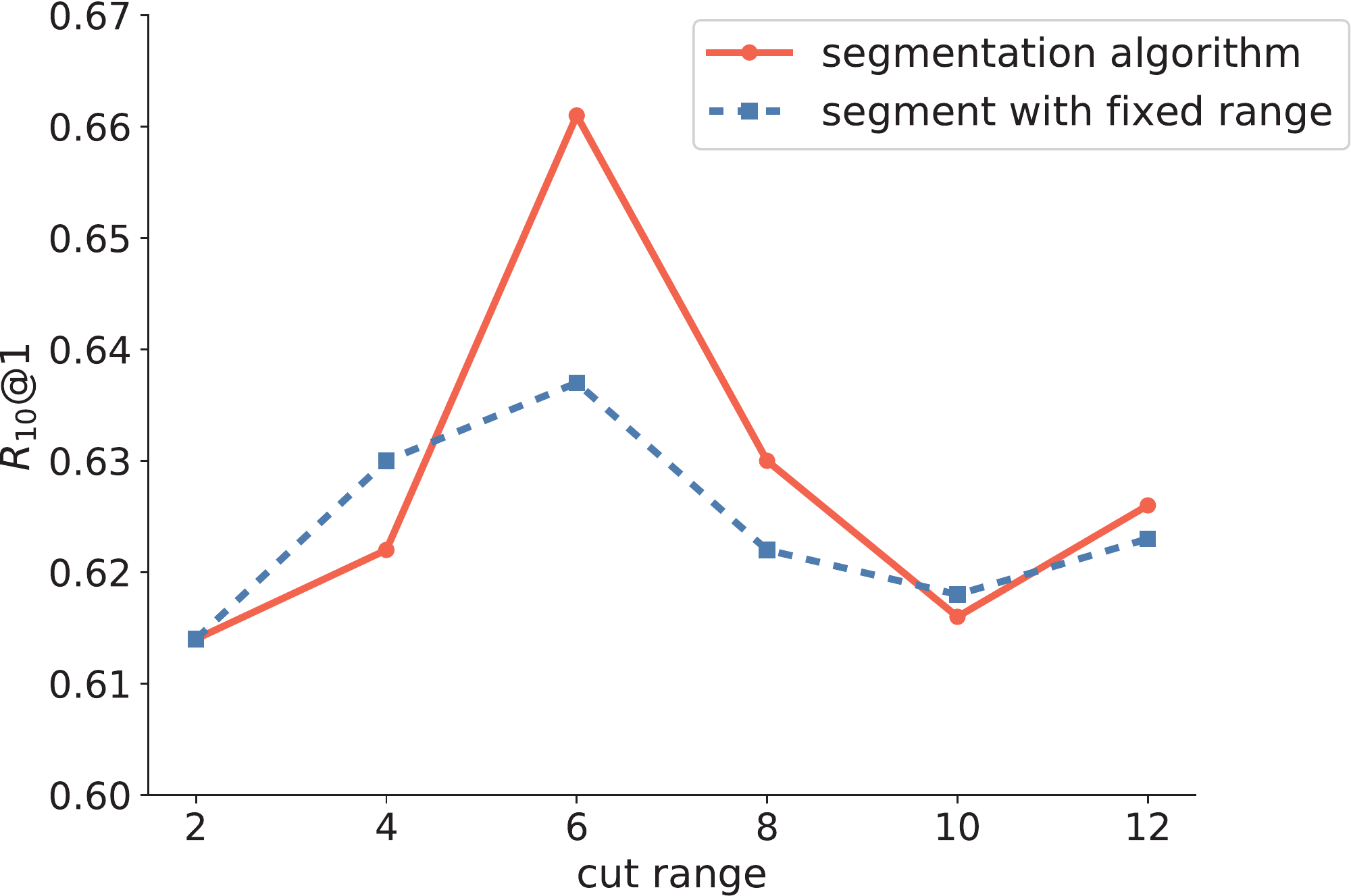}
\caption{$R_{10}@1$ for different ranges on E-commerce.}
\label{fig: fixed_interval_E}
\end{figure}

In order to investigate the effectiveness of our segmentation algorithm in the response selection task, we use a simple method: just segment using fixed ranges 2, 4, 6, 8, 10. Besides, we also apply our segmentation algorithm with the corresponding range $R$. Results are shown in Figure \ref{fig: fixed_interval_E}.

For both methods, with the increase of cut range, range of 6 performs best. Both too small and large intervals hurt performance. Besides, applying our segmentation algorithm performs better than just using fixed ranges in most ranges especially in range of 6, which shows the effectiveness of our proposed topic-aware segmentation algorithm.
\section{Conclusion}
This paper presents the first topic-aware multi-turn dialogue modeling design in terms of explicitly segmenting and extracting topic-aware context utterances for retrieval-based dialogue systems. To fulfill our research purpose, we build two new datasets with topic boundary annotation and propose an effective topic-aware segmentation algorithm as prerequisites of this work. Then, we propose topic-aware dual-attention matching network to further improve the matching of response and segment contexts. Our models are evaluated on three benchmark datasets, showing new state-of-the-art, which verifies that the topic-aware clues and the related model design are effective.

\bibliography{aaai21}
\end{document}